# An Improved Image Mining Technique For Brain Tumour Classification Using Efficient classifier

P.Rajendran
Department of Computer science and Engineering
K. S. Rangasamy College of Technology,
Tiruchengode-637215, Tamilnadu, India.
Phone: +91 4288 274741,
Fax: +91 4288 274757
.

M.Madheswaran
Center for Advanced Research, Department of Electronics
and Communication Engineering,
Muthayammal Engineering College,
Rasipuram – 637 408, Tamilnadu, India.
Phone: +91 4287 220837, Fax: +91 4287 226537
.

*Abstract*— *An improved image mining technique for brain tumor classification using pruned association rule with MARI algorithm is presented in this paper. The method proposed makes use of association rule mining technique to classify the CT scan brain images into three categories namely normal, benign and malign. It combines the low-level features extracted from images and high level knowledge from specialists. The developed algorithm can assist the physicians for efficient classification with multiple keywords per image to improve the accuracy. The experimental result on pre-diagnosed database of brain images showed 96% and 93% sensitivity and accuracy respectively.*

*Keywords-Data mining; Image ming; Association rule mining; Medical Imaging; Medical image diagnosis;. Classification.*

## I. INTRODUCTION

In health care centers and hospitals, millions of medical images have been generated daily. Analyses have been done manually with an increasing number of images. Even after analyzing a minimal number of images, radiologist becomes more tiresome. Nowadays, physicians are providing with computational techniques in assisting the diagnosis process. In the recent past, the development of Computer Aided Diagnosis (CAD) systems for assisting the physicians for making better decisions have been the area of interest [1]. This has motivated the research in creating vast amount of image database. In CAD method, computer output has been used as a second opinion for radiologist to diagnose the information more confident and quicker mechanism as compared to manual diagnosis.

Pathologies are clearly identified using automated CAD system [2]. It also helps the radiologist in analyzing the digital images to bring out the possible outcomes of the diseases. In the last few years, inexpensive and available means of database containing rich medical data have been provided through the internet for health services globally. It has been reported that the brain tumor is one of the major cause leading to higher incidence of death in human. Physicians have faced a challenging task in extracting the features and decision making. The Computerized Tomography (CT) has been found to be the most reliable method for early detection of tumors because this modality is the most used in radiotherapy planning for two main reasons.

The first reason is that scanner images contain anatomical information which offers the possibility to plan the direction and the entry points of the radiotherapy rays which have to target the tumor and to avoid some risk organs. The second reason is that CT scan images are obtained using rays, which is the same physical principle as radiotherapy. This is very important because the radiotherapy rays intensity can be computed from the scanner image intensities. Due to the high volume of CT [3] images to be used by the physicians, the accuracy of decision making tends to decrease. This has further increased the demand to improve the automatic digital reading for decision making [4]. It also significantly improves in the field of conservative treatment of CAD diagnosis. It is an interdisciplinary field that combines techniques like data mining, digital image processing, radiology and usability among others.

In this paper image mining concepts have been used. It deals with the implicit knowledge extraction, image data relationship and other patterns which are not explicitly stored in the images. This technique is an extension of data mining to image domain. It is an inter disciplinary field that combines techniques like computer vision, image processing, data mining, machine learning, data base and artificial intelligence [5]. The objective of the mining is to generate all significant patterns without prior knowledge of the patterns [6]. Rule mining has been applied to large image data bases [7]. Mining has been done based on the combined collections of images and it is associated data. The essential component in image mining is the identification of similar objects in different images [8].

The method proposed in this paper classifies the brain CT scan images into three categories: normal, benign and malignant. Normal ones are those characterizing a healthy patient, benign cases represents CT scan brain images showing a tumor that are not formed by cancerous cells, and Malign cases are those brain images that are taken from patients with cancerous tumors. CT scan brain images are among the most difficult medical images to be read due to their low contrast and differences in the type of tissues. This paper illustrates the importance of data cleaning phase in building an accurate data mining architecture for image classification [9, 10, and 11].

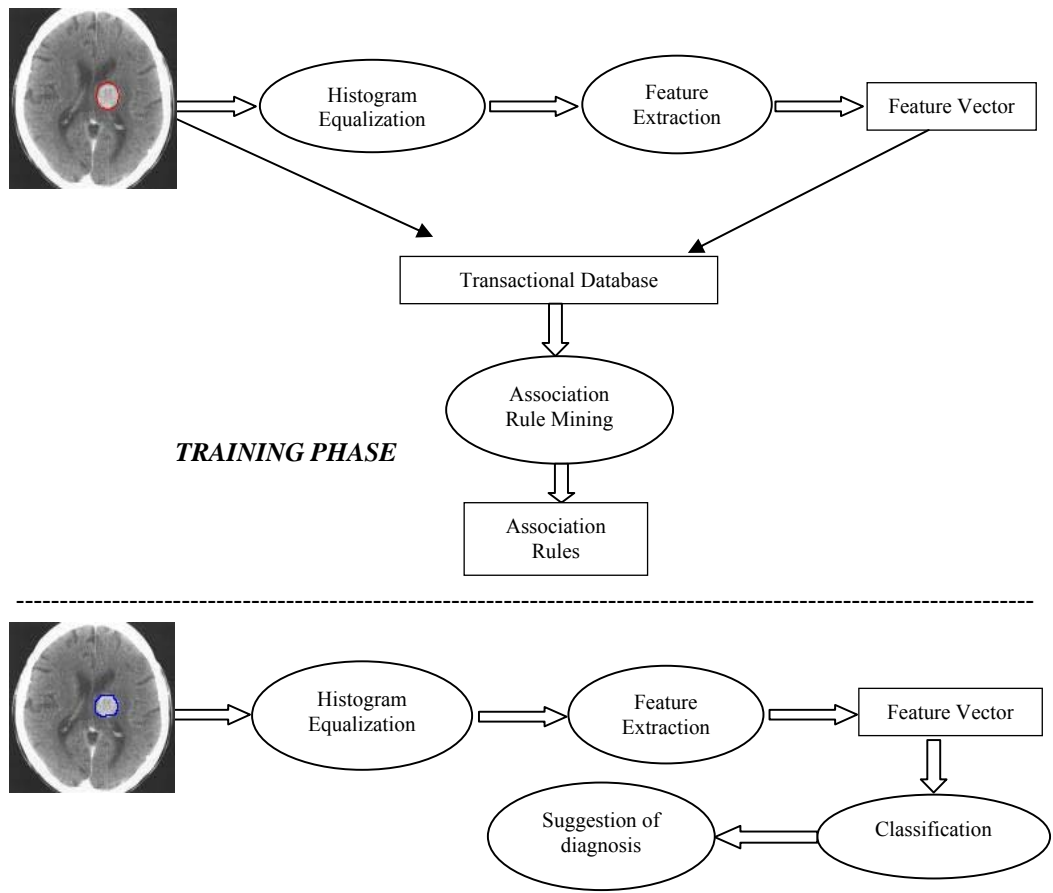

Figure 1. Overview of the proposed system

The method presented here is based on the associative classification scheme. This approach has an advantage of selecting only the most relevant features during mining process and obtaining multiple keywords when processing a test image [12].

## II. SYSTEM DESCRIPTION

Overview of the proposed system is shown in Fig 1. The proposed system is mainly divided into two phases: the training phase and the test phase. Data cleaning and feature extraction are common for both the training set of brain images and the test set [13, 14]. In the training phase, features are extracted from the images, represented in the form of feature vectors. Next, the features are discretized into intervals and the processed feature vector is merged with the keywords related with the training images [15]. This transaction representation is submitted to the MARI (Mining Association Rule in Image database) algorithm for association rule mining, which finally produces a pruned set of rules representing the actual classifier [16, 17]. In the test phase, the feature vector obtained from the test images are submitted to the classifier which makes use of the association rules to generate keywords to compose the diagnosis of the test image. These keywords have been used to classify the three categories of CT scan brain images as normal image, benign (tumor without cancerous tissues) image and malignant (tumor with cancerous tissues) image.

### A. Pre-Processing

Since most of the real life data is noisy, inconsistent and incomplete, preprocessing becomes necessary [18]. The cropping operation can be performed to remove the background, and image enhancement can be done to increase the dynamic range of chosen features so that they can be detected easily. In general, most of the soft tissues have overlapping gray-levels and the condition of illumination at that time of CT scan taken is also different. The histogram equalization can be used to enhance the contrast within the soft tissue of the brain images and also hybrid median filtering technique can be used to improve the image quality. Good texture feature extraction can be done by increasing the dynamic range of gray-levels using the above mentioned technique [19].

### B. Texture Feature Extraction

As the tissues present in brain are difficult to classify using shape or intensity level of information, the texture feature extraction is found to be very important for further classification [20,21, and 22]. The analysis and characterization of texture present in the medical images can be done using several approaches like run length encoding, fractal dimension, and discrete wavelet transform and co-occurrence matrices [23, 24].

Though many texture features have been used in the medical image classification, Spatial Gray Level Dependent Features (SGLDF) can be used to calculate the intersample distance for better diagnosis [25, 26]. In order to detect the abnormalities in medical images association rule mining is built using texture information [27, 28]. This information can be categorized by the spatial arrangement of pixel intensities. In order to capture the spatial distribution of the gray levels within the neighborhood, two dimensional co-occurrence matrices can be applied to calculate the global level features and pixel level features.

The following ten descriptors can be used for extracting texture features.

$$\text{Entropy} = -\sum_{i}^{M}\sum_{j}^{N} P[i,j] \log P[i,j] \quad (1)$$

$$\text{Energy} = \sum_{i}^{M}\sum_{j}^{N} P^{2}[i,j] \quad (2)$$

$$\text{Contrast} = \sum_{i}^{M}\sum_{j}^{N} (i-j)^{2} P[i,j] \quad (3)$$

$$\text{Homogeneity} = \sum_{i}^{M}\sum_{j}^{N} \frac{P[i,j]}{1+|i-j|} \quad (4)$$

$$\text{SumMean} = \frac{1}{2}\sum_{i}^{M}\sum_{j}^{N} (i*P[i,j] + j*P[i,j]) \quad (5)$$

$$\text{Variance} = \frac{1}{2}\sum_{i}^{M}\sum_{j}^{N} ((i-\mu)^{2} P[i,j] + (j-\mu)^{2} P[i,j]) \quad (6)$$

$$\text{Maximum\_Probability} = \underset{i,j}{\text{Max}}^{M,N} P[i,j] \quad (7)$$

$$\text{Inverse\_Diference\_Moment} = \sum_{i}^{M}\sum_{j}^{N} \frac{P[i,j]}{|i-j|^{k}} \quad (8)$$

$$\text{Cluster\_Tendency} = \sum_{i}^{M}\sum_{j}^{N} (i+j-2\mu)^{k} P[i,j] \quad (9)$$

$$\text{Correlation} = \sum_{i}^{M}\sum_{j}^{N} \frac{(i-\mu)(j-\mu) P[i,j]}{\sigma^{2}} \quad (10)$$

where P is the normalized co-occurrence matrix, (i, j) is the pair of gray level intensities and M by N is the size of the co-occurrence matrix. The intersample distance is estimated based on estimation of the second-order joint conditional probability density function for the pixel (i, j), $P[i,j | d,\theta]$ for $\theta = 0°, 45°, 90°$ and $135°$. The function is the probability that two pixels which are located with an inter sample distance d and a direction $\theta$. The estimated joint conditional probability density functions are defined as

$$P[i,j | d, 0°] = \# \left\{ \begin{array}{l} ((k,l),(m,n)) \in [L_x \times L_y] \times [L_x \times L_y]: \\ k = m, |l-n| = d, S(k,l) = i, S(m,n) = j \end{array} \right\} / T(d, 0°)$$

$$P[i,j | d, 45°] = \# \left\{ \begin{array}{l} ((k,l),(m,n)) \in [L_x \times L_y] \times [L_x \times L_y]: \\ (k-m = d, l-n = -d) \text{ or} \\ (k-m = -d, l-n = d) \\ S(k,l) = i, S(m,n) = j \end{array} \right\} / T(d, 45°)$$

$$P[i,j | d, 90°] = \# \left\{ \begin{array}{l} ((k,l),(m,n)) \in [L_x \times L_y] \times [L_x \times L_y]: \\ (k-m = d, l = n), S(k,l) = i, \\ S(m,n) = j \end{array} \right\} / T(d, 90°)$$

$$P[i,j | d, 135°] = \# \left\{ \begin{array}{l} ((k,l),(m,n)) \in [L_x \times L_y] \times [L_x \times L_y]: \\ (|k-m| = d, l-n = -d), \\ S(k,l) = i, S(m,n) = j \end{array} \right\} / T(d, 135°)$$

where # denotes the number of elements in the set, S(x, y) is the image intensity at the point (x, y) and $T(d, \theta)$ stands for the total number of pixel pairs within the image which has the intersample distance d and direction $\theta$.

Co-occurrence matrices can be calculated for the directions $0°, 45°, 90°, 135°$ and their respective pixels are denoted as 1, 2, 3 and 4. Once the co-occurrence matrix is calculated around each pixel, the features such as entropy, energy, variance, homogeneity and inverse variance can be obtained for each matrix with respect to the intersample distance. From the co-occurrence matrices the feature vectors can be calculated and stored in the transaction database. Next, the continuous valued features are discretized into intervals, where each interval represents an item in the process of mining association rules [15].

### III. BUILDING THE CLASSIFIER

The extracted features can be stored in to transaction database for classification using the trained brain images. The image mining techniques can be applied to match the extracted features with trained sets for proper classification [29].

#### A. Association Rule Mining

Association rule mining aims at discovering the associations between items in a transactional database [8, 27, and 30]. Given a set of transactions D= {t1, t2….tn} and a set of items I= {i1, i2 ,…in} such that the transaction T in D is a set of items in I, an association rule is an implication of the form $X \Rightarrow Y$, where X is called body or antecedent of the rule, and Y is called as head or consequent of the rule [31]. The problem of mining association rules involves finding rules that satisfy minimum support and minimum confidence specified by the user. In this approach, modified version of ARC-AC algorithm [32] can be used for mining the association among the features from the transactional database. The proposed algorithm named MARI has been explained as follows.

**Algorithm:** MARI Find association on the training set of the transactional database.

**Input:** A set of Image patches (P1) of the form $P_i : \{k_1, k_2 \ldots k_m, f_1, f_2, \ldots f_n\}$ where $k_i$ is a keyword attached to the patches and $f_j$ are the selected features for the patches, a minimum support threshold $\sigma$.

**Output:** A set of association rules of the form $f_1 \wedge f_2 \wedge \ldots \ldots \wedge f_n \Rightarrow k_i$ where $k_i$ is a keyword, $f_j$ is a feature and kw is a class category.

**Method:**
(1)  $C_0 \leftarrow$ {candidate keywords and their support}
(2)  $F_0 \leftarrow$ {frequent keywords and their support}
(3)  $C_1 \leftarrow$ {candidate keyword 1 item sets and their support}
(4)  $F_1 \leftarrow$ {frequent 1 item sets and their support}
(5)  $C_2 \leftarrow$ {candidate pairs $(k, f)$, such that $(k, f) \in P_1$ and $k \in F_0$ and $f \in F_1$}
(6)  For each patches p in $P_1$ do {
(7)      For each kw = $(k, w)$ in $C_2$ do {
(8)          kw.support $\leftarrow$ kw.support. count $(kw, p)$
(9)      }
(10)  }
(11) $F_2 \leftarrow \{kw \in C_2 | kw.support > \sigma\}$
(12) $P_2 \leftarrow$ Filter Table $(P_1, F_2)$
(13) For (i $\leftarrow$ 3; $F_{i-1} \neq \phi$; i $\leftarrow$ i+1) do {
(14)     $C_i \leftarrow (F_{i-1} \bowtie F_2)$
(15)     $C_i \leftarrow C_i - \{kw | (i-1) \text{Item-set of } kw \notin F_{i-1}\}$
(16)     $P_i \leftarrow$ Filter Table $(P_{i-1}, F_{i-1})$
(17)         For each Patches p in $P_i$ do {
(18)     for each kw in $C_i$ do
(19)         {
(20)     kw.support $\leftarrow$ kw.support + count (kw, p)
(21)         }
(22)     }
(23)     $F_i \leftarrow \{kw \in C_i | kw.support > \sigma\}$
(24)     }
(25) }
(24) sets $\leftarrow \bigcup_i \{kw \in F_i | i > 1\}$
(25) Rule=$\phi$
(26) for each item set I in sets do {
(27) Rule$\leftarrow$Rule+$\{f \Rightarrow kw | f \cup kw \in I \wedge f$ Is an itemset $\wedge kw \in C_0\}$
(28) }

The association rules are constrained such that the antecedent of the rules is composed of conjunction of features from the brain image while the consequent of the rule is always the class label to which the brain image belongs [33].

#### B. Pruning Techniques

The rules generated in the mining phase are expected to be very large. This could be a problem in applications where fast responses are required. Hence, the pruning techniques become necessary to eliminate the specific rules and which are conflicting with the same characteristics pointing different categories [34, 35].

This can be achieved using the following conditions:

Condition 1: Given two rules $R1 \Rightarrow C$ and $R2 \Rightarrow C$, the first rule is a general rule if $R1 \subseteq R2$. To attain this, ordering the association rules must be done as per condition 2.

Condition 2: Given two rules R1 and R2, R1 is higher ranked than R2 if:
(1) R1 has higher confidence than R2,
(2) If the confidences are equal, support of R1 must exceed support of R2.

(3) If both confidences and supports are equal, but R1 has less attributes in left hand side than R2.

Condition 3: The rules $R1 \Rightarrow C1$ and $R1 \Rightarrow C2$, represents are conflict in nature. Based on the above conditions, duplicates have been eliminated. The set of rules that are selected after pruning represents the actual classifier. These conditions have been used to predict to which class the new test image belongs.

### C. Classification Of Test Image

After the completion of training phase, an actual classifier with pruned set of association rules can be built for training the brain images [36]. Each training image is associated with a set of keywords. Keywords are representative words chosen by a specialist to use in the diagnosis of a medical image. The knowledge of specialists should also be considered during the processing of mining medical images in order to validate the results. The extracted features of the test image and the feature vector generated can be submitted to the classifier, which uses the association rules and generates set of keywords to compose the diagnosis of a test image.

Algorithm:
Input: Feature vector F of the test image, threshold
Output: set of keywords S
Method:
(1)      for each rule $r \in R$ of the form body $\rightarrow$ head do
(2)      {
(3)           for each itemset $h \in$ head do
(4)           {
(5)      if body matches F then
(6)           increase the number of matches by 1
(7)           Else
(8)      increase the number of non matches by 1
(9)           }
(10)    }
    // to generate keywords
(11) for each rule $r \in R$ of the form body $\rightarrow$ head do
(12) {
(13)    for each item set $h \in$ R head do
(14)         {
(15)    if $(n(Mh)/n(Mh) + n(Nh))) \geq T$ then
(16)    if $h \notin S$ then
(17)    add h in S
(18)         }
(19) }
(20)    return S

This classifier returns the multiple classes when processing a test image. The algorithm developed has been employed to generate suggestions for diagnosis. This algorithm stores all item sets (i.e. Set of keywords) belonging to the head of the rules in a data structure. An item set h is returned in the suggested diagnosis if the condition is satisfied as the given equation

$$n(M_h)/(n(M_h)+n(N_h)) \geq T \qquad (15)$$

where, $n(M_h)$ is the number of matches of the item set h and $n(N_h)$ is the number of non-matches. Threshold T is employed to limit the minimal number of matches required to return an item set in the suggested diagnosis. A match occurs when the image features satisfy the body part of the rule.

### D. Performance evaluation criteria

The confusion matrix can be used to determine the performance of the proposed method and is shown in Fig 2. This matrix describes all possible outcomes of a prediction results in table structure. The possible outcomes of a two class prediction be represented as True positive (TP), True negative (TN), False Positive (FP) and False Negative (FN). The normal and abnormal images are correctly classified as True Positive and True Negative respectively. A False Positive is when the outcome is incorrectly classified as positive (yes) when it is a negative (no). False Positive is the False alarm in the classification process. A false negative is when the outcome is incorrectly predicted as negative when it should have been in fact positive.

From the confusion matrix, the precision and recall values can be measured using the formula. Precision: It is defined as the fraction of the classified image, which is relevant to the predictions. It is represented as

$$\mathrm{Precision} = \frac{TP}{TP+FP} \qquad (16)$$

Recall: It is defined as the fraction of the classified image for all the relevant predictions. It is given as

$$\mathrm{Recall} = \frac{TP}{TP+FN} \qquad (17)$$

|        |     | Identified |     |
|--------|-----|------------|-----|
|        |     | Yes        | No  |
| Actual | Yes | TP         | FN  |
|        | No  | FP         | TN  |

Figure 2. Confusion matrix

## IV. RESULT AND DISCUSSIONS

An experiment has been conducted on a CT scan brain image data set based on the proposed flow diagram as shown in Fig 1. The pre-diagnosed databases prepared by physicians are considered for decision making. Fig 3(a) represents the original input image and Fig 3(b) shows the result of histogram equalization and hybrid median filtered original image, which is used to reduce the different illumination conditions and noises at the scanning phase.

After preprocessing, feature extraction has been done to remove the irrelevant and redundant content of the information present in the input image [23]. Haralick co-occurrence method has been used to determine the discrimination of the tissue level variations and shown in Fig.4. In Fig 4(a) pixel 1 represents $0°$, pixel 2 represents $45°$, pixel 3 represents $90°$ and pixel 4 represents $135°$ from the centered pixel, at a distance value of one. Fig 4(b) shows the preprocessed CT scan brain image merged with angle representation. Fig 4(c) represents the pixel representation matrix for distance one and degree zero, similarly the pixel representation matrix have to be calculated for the remaining degrees. From the pixel representation matrix, the co-occurrence matrices are also calculated and represented in the Fig 4(d).

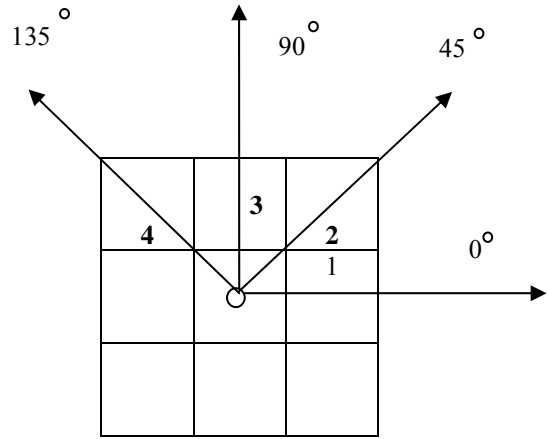

Figure 4(a). Matrix representation for center pixel and all around pixels

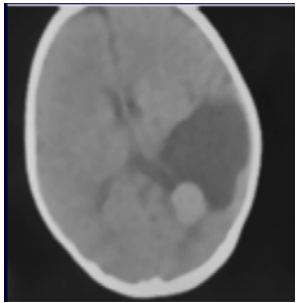

Figure 3(a). Input CT scan brain image

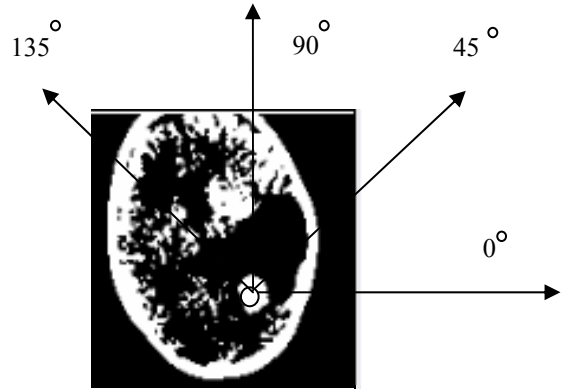

Figure 4(b). Preprocessed CT scan brain image merged with angle representation

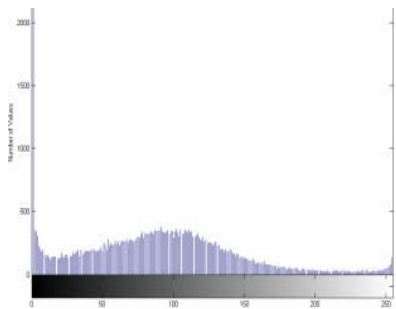

Figure 3(b). Histogram equalized image

| 0 | 0 | 1 | 3 | 1 |
|---|---|---|---|---|
| 3 | 1 | 1 | 3 | 1 |
| 2 | 1 | 3 | 0 | 3 |
| 3 | 2 | 1 | 0 | 3 |
| 3 | 3 | 2 | 1 | 2 |

Figure 4(c). Pixel representation matrix for Zero Degree

| i /j | 0 | 1 | 2 | 3 |
|---|---|---|---|---|
| 0 | 1 | 1 | 0 | 2 |
| 1 | 1 | 1 | 1 | 3 |
| 2 | 0 | 3 | 0 | 0 |
| 3 | 1 | 3 | 2 | 1 |

Figure 4(d). Co-occurrence matrix for distance one and Degree zero

Fig 5 represents the flow of texture feature extraction and object segregation process. For each object co-occurrence matrix has been calculated and texture features are extracted. Four different directions $0°, 45°, 90°, 135°$ generate 16 co-occurrence matrices. The texture features are then calculated for each co-occurrence matrix and stored in the database. The feature vectors have calculated and the obtained vectors are stored in the transaction database from the co-occurrence matrix values.

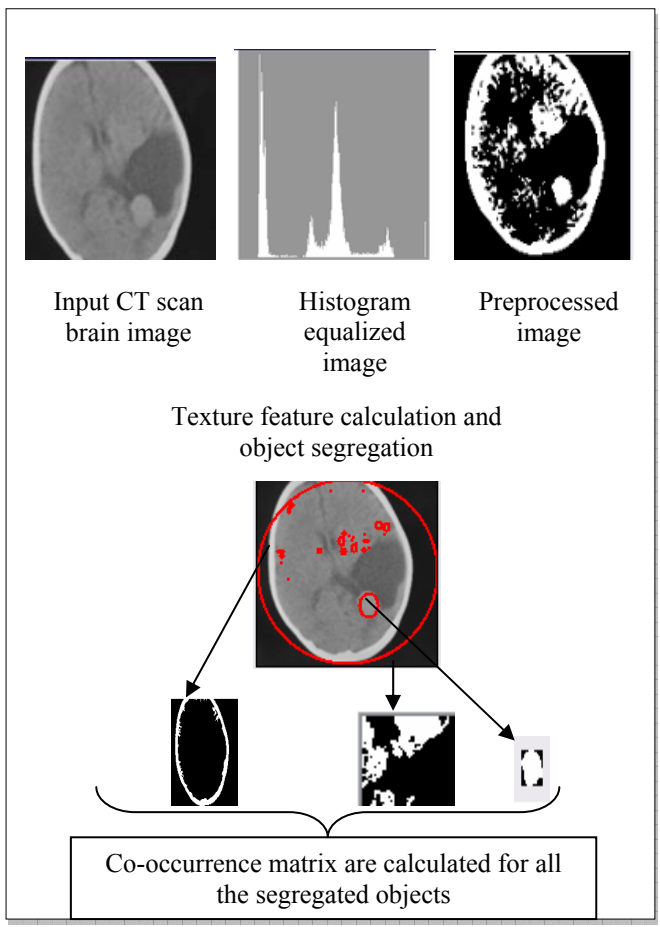

Figure 5. Texture feature extraction and object segregation

| Image | Diagnosis |
|---|---|
|  | Malign<br>Assessment=5<br>Subtlety=5<br>Abnormality=1 |
|  | Benign<br>Assessment=3<br>Subtlety=4<br>Abnormality=1 |
|  | Benign<br>Assessment=3<br>Subtlety=3<br>Abnormality=1 |

Figure 6. Images of the dataset and their diagnoses

MARI algorithm has been applied on the transaction database which consists of the feature vectors and the diagnosis information about the training CT scan images. Fig 6 represents the images of the sample dataset and their diagnosis information.

In Fig 7 the precision and recall values of the proposed method, Association Rule Mining (ARM) method and Naïve Bayesian method are plotted in the graph. It shows that the performance of proposed method is better compared to the existing methods.

The effectiveness of the proposed method has been estimated using the following measures:

Accuracy= (TP+TN)/ (TP+TN+FP+FN)

Sensitivity= TP/ (TP+FN)

Specificity= TN/ (TN+FP)

where, TP, TN, FP, and FN are the number of True Positive cases (abnormal cases correctly classified), the number of True Negatives (normal cases correctly classified), the number of False Positives (normal cases classified as abnormal), and the number of False Negatives (abnormal cases classified as normal) respectively. Accuracy is the proportion of correctly diagnosed cases from the total number of cases. Sensitivity measures the ability of the proposed method to identify abnormal cases. Specificity measures the ability of the method to identify normal cases. The value of minimum confidence is set to 97% and the value of minimum support is set to 10%. The features of the test images and the association rules have been generated using the threshold value=0.001. The results show that the proposed classifier gives higher values of sensitivity, specificity and accuracy such as 96%, 90% and 93% respectively. In order to validate the obtained results, the algorithmic approach has been compared with the well known classifier, a naive bayesian classifier and associative classifier [37, 38 and 39].

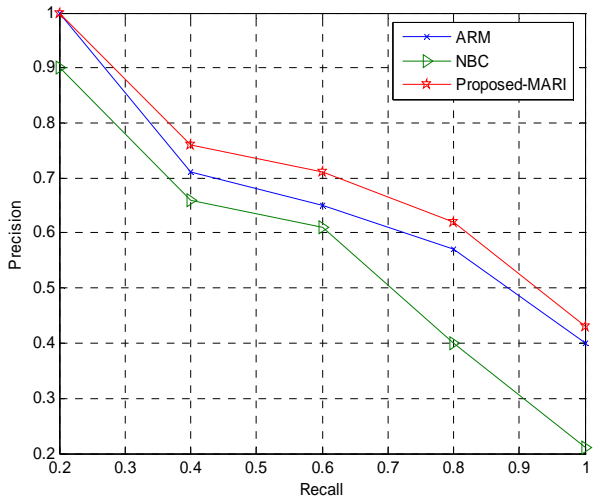

Figure 7.  P & R graph using naive bayesian, association rule mining and MARI association rule mining

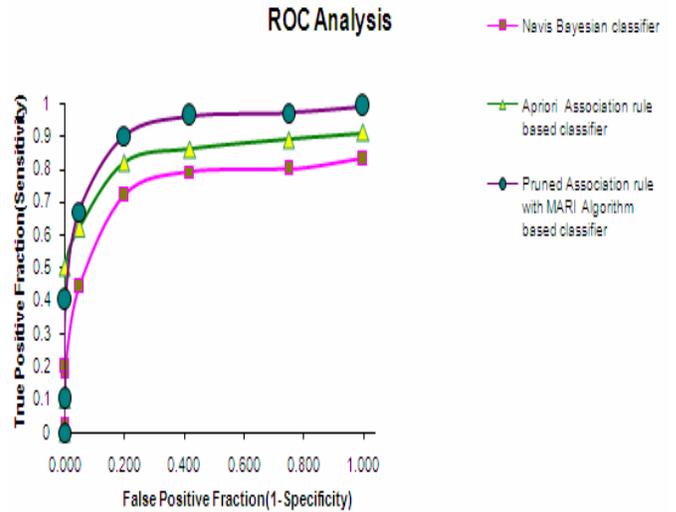

Figure 8. Tumor classification by ROC analysis

Table 1 illustrates the sensitivity, accuracy, specificity, area under the curve (Az), standard error (SE) and execution time of naive bayesian classifier, association rule mining and proposed method. The experimental results have shown that the proposed method achieves high sensitivity (up to 96%), accuracy (up to 93%) and less execution time and standard error in the task of support decision making system.

Table 2 represents the results of the classifiers, here 150 images are taken for training and 95 images are taken for the testing in both benign and malignant categories, which are classified using different classifiers. The results show that the proposed system gives better percentage of correct classification as compared to naive bayesian classifier and association rule based classifier.

TABLE 1 PERFORMANCE COMPARISION FOR CLASSIFIERS

| Classes | No. of data Training/ Testing | No. of correctly classified data | | | Percentage of correct classification | | |
|---|---|---|---|---|---|---|---|
| | | Naive Bayesian classifier | Association rule based classifier | Pruned Association rule with MARI Algorithm based classifier | Naive Bayesian classifier | Association rule based classifier | Pruned Association rule with MARI Algorithm based classifier |
| Benign | 150 / 95 | 84 | 91 | 93 | 88.4 | 95.78 | 97.90 |
| Malignant | 150 / 95 | 85 | 92 | 94 | 89.4 | 96.84 | 98.95 |
| Average | | | | | 89.9 | 96.31 | 98.42 |

TABLE 2  RESULTS OF THE CLASSIFIERS

| Approach | Sensitivity (%) | Specificity (%) | Accuracy(%) | $A_z$ | SE | Time (ms) |
|---|---|---|---|---|---|---|
| Naive Bayesian classifier [22] | 75 | 63 | 70 | 0.89 | 0.08 | 30.91 |
| Association rule based classifier [22] | 95 | 84 | 91 | 0.91 | 0.03 | 9.75 |
| Pruned Association rule with MARI Algorithm based classifier | **96** | **90** | **93** | 0.98 | 0.02 | 2.15 |

The Receiver Operating Characteristic (ROC) curves are plotted with respect to sensitivity and specificity. The area under the ROC plays a vital role since it has been used to determine the overall classification accuracy. Fig 8 shows the comparison of the ROC curve for various classifiers. It clearly shows that the proposed mining based classification with pruned rules has higher value of $(A_z)$ as compared to other methods.

## V. CONCLUSION

An improved image mining technique for brain tumor classification using pruned association rule with MARI algorithm has been developed and the performance is evaluated. The proposed algorithm has been found to be performing well compared to the existing classifiers. The accuracy of 93% and sensitivity of 96% were found in classification of brain tumors. The developed brain tumor classification system is expected to provide valuable diagnosis techniques for the physicians.

ACKNOWLEDGEMENT

The authors would like to express their gratitude to Dr. D. Elangovan, Pandima CT scan centre, Dindigul for providing the necessary images for this study.

REFERENCES

[1] X.R. Marcela, H.B. Pedro, T.J. Caetano, M.A.M. Paulo, A.R. Natalia, and J.M Agma , "Supporting Content-Based Image Retrieval and Computer-Aided Diagnosis Systems with Association Rule-Based Techniques," Data & Knowledge Engineering, in press.

[2] G.D. Tourassi, "Journey toward Computer-Aided diagnosis role of image texture analysis," Radiology, 1999, pp 317-320.

[3] S.R. Daniela, "Mining Knowledge in Computer Tomography Image Databases," Multimedia Data Mining and Knowledge Discovery, Springer London, 2007.

[4] B. Thomas, "Analyzing and mining image databases. Reviews Drug Discovery Today," Biosilico 10(11), 2005, pp.795-802.

[5] C. Ordonez, E. Omiecinski ,"Image mining: A new approach for data mining," Technical Report GITCC-98-12, Georgia Institute of Technology, College of Computing, 1998, pp 1-21.

[6] H. Wynne, L.L Mong ,and J. Zhang, "Image mining: trends and developments. Journal of Intelligent Information Systems," 19 (1): 2002, pp 7–23.

[7] P. Stanchev , M. Flint, "Using Image Mining For Image Retrieval," In. Proc: IASTED conf. Computer Science and Technology, 2003, pp. 214-218.

[8] C. Ordonez, E.Omiecinski, "Discovering association rules based on image content," In Proc: IEEE Forum ADL, 1999, pp. 38–49.

[9] V. Megalooikonomou, J. For ,F. Makedon , "Data Mining in Brain Imaging," Statistical Methods in Medical Research, 2000, pp. 359-394.

[10] H. Pan, L.Jianzhong, W.Zhang, "Incorporating domain knowledge into medical image clustering," Applied Mathematics and Computation, 2007, pp. 844–856.

[11] J.B.W. Pluim, J.B.A. Maintz, and M.A. Viergever, "Mutual-information-based registration of medical images," A survey. IEEE Transactions on Medical Imaging, 22(8):2003, pp. 986-1004.

[12] M.L. Antonie, O.R. Zaiane, and A.Coman, "Associative classifiers for medical images," Revised Papers from MDM/KDD and PAKDD/KDMCD, 2002, 68–83.

[13] B.A. Dogu, H. Markus, A.Tuukka, D. Prasun, and H. Jari , "Texture Based Classification and Segmentation of Tissues Using DT-CWT Feature Extraction Methods," In Proc: 21st IEEE International Symposium on Computer-Based Medical Systems, 2008, pp.614-619.

[14] P. Dollar, T. Zhuowen, T. Hai, and S. Belongie, " Feature Mining for Image Classification," In Proc: IEEE Conference on Computer Vision and Pattern Recognition, 2007, pp. 1-6.

[15] J. Dougherty, R. Kohavi, and M. sahami, "Supervised and Unsupervised Discretization of continuous features," In Proc: 12th International Conference Machine Learning, 1995, pp.56-69.

[16] R. Agrawal, T. Imielinski, And A.N. Swami, "Mining association rules between sets of items in large databases," In Proc: ACMSIGMOD Int. Conf. Manage, Washington, DC, 1993, pp. 207-216.

[17] S. Kotsiantis, D.Kanellopoulos, "Association Rules Mining: A Recent Overview," GESTS International Transactions on Computer Science and Engineering, 32 (1):2006, pp. 71-82.

[18] R. Agrawal, R.Srikant, "Fast algorithms for mining association rules," In Proc: Int. Conf. VLDB, Santiago, Chile, 1994, pp. 487-499.

[19] J.C. Felipe, A.J.M Traina, and C. Traina, "Retrieval by content of medical images using texture for tissue identification," In Proc: 16th IEEE Symp. Computer-Based Med. Systems. CBMS 2003, pp. 175–180.

[20] R. Abraham, J.B. Simha,and S.S Iyengar, "Medical datamining with a new algorithm for Feature Selection and Naive Bayesian classifier.," In Proc: 10th International Conference on Information Technology (ICIT), 2007, pp. 44-49.

[21] C. Christophe, S.G. Jean, L.M. Gael, and K. Michel, "Efficient Data Structures and Parallel Algorithms for Association Rules Discovery," In Proc: Fifth Mexican International Conference in Computer Science (ENC), 2004, pp. 399-406.

[22] P.G. Foschi, D.Kolippakkam, H. Liu, and A. Mandvikar , " Feature extraction for image mining," In Proc: 8th Int. Workshop Multimedia Inf. Syst, Tempe, AZ, 2002, pp. 103-109.

[23] R.M. Haralick, K.Shanmugam, and I. Distein, "Textural features for image classification.," IEEE Trans. Syst, Man, Cybern, vol. SMC-3, 1973, pp. 610–621.

[24] A. Ranjit, B.S. Jay, and S.S. Iyengar, "Medical Data mining with a New Algorithm for Feature Selection and Naive Bayesian Classifier," In Proc: 10th International Conference on Information Technology (ICIT), 2007, pp.44-49.

[25] L. Hui, W. Hanhu, C. Mei, and W. Ten , "Clustering Ensemble Technique Applied in the Discovery and Diagnosis of Brain Lesions," In Proc: Sixth International Conference on Intelligent Systems Design and Applications (ISDA) , vol. 2: 2006, pp. 512-520.

[26] C.F. Joaquim, X.R. Marcela, P.M.S. Elaine, J.M.T. Agma, and T.J. Caetano, "Effective shape-based retrieval and classification of mammograms," In Proc: ACM symposium on Applied computing, 2006, pp. 250 – 255.

[27] N.R. Mudigonda, R.M. Rangayyan , "Detection of breast masses in mammograms by density slicing and texture flow-field analysis," IEEE Trans. Med. Imag. 20(12), 2001, pp. 1215–1227.

[28] K. Murat, I.M. Cevdet,"An expert system for detection of breast cancer based on association rules and neural network," An International Journal Expert Systems with Applications 36: 2009, pp. 3465–3469.

[29] K. Lukasz, W. Krzysztof, "Image Classification with Customized Associative Classifiers," In Proc: International Multiconference on Computer Science and Information Technology, 2006, pp. 85–91.

[30] E. Laila, A.A. Walid, "Mining Medical Databases using Proposed Incremental Association Rules Algorithm (PIA)," In Proc: IEEE Second International Conference on the Digital Society, 2008, pp 88-92.

[31] A. Olukunle, S.A. Ehikioya, "A Fast Algorithm for Mining Association Rules in Medical Image Data," In Proc: IEEE Canadian Conf. Electr. Comput. Eng. Conf, 2002, pp. 1181–1187.

[32] L.A. Maria, R.Z. Osmar, and C. Alexandru, "Associative Classifiers for Medical Images.Mining," Lecture Notes in Computer Science, Multimedia and Complex Data, Springer Berlin / Heidelberg, 2003, pp.68-83.

[33] X.Wang, M.Smith, and R. Rangayyan, " Mammographic information analysis through association-rule mining," In Proc: IEEE CCGEI, 2004, pp. 1495-1498.

[34] B. Liu, W. Hsu, Y. Ma, " Pruning and Summarizing the Discovered Associations," In Proc: ACM SIGKDD International Conference on Knowledge Discovery & Data Mining , 1999, pp. 81-105.

[35] R.Z. Osmar, L.A. Maria," On Pruning and Tuning Rules for Associative Classifiers," In Proc: 9th International Conference(KES) ,K-Based-Intelligent Information and Engineering Systems Melbourne, part III, 2005.

[36] S. T. Vincent, W. Ming-Hsiang, and S.J. Hwung, "A New Method for Image Classification by Using Multilevel Association Rules," In Proc: 21st International Conference on Data Engineering Workshops (ICDEW), 2005, pp.1180-1188.

[37] G.H. John, P. Langley, "Estimating Continuous Distributions in Bayesian Classifiers," In Proc: 11th conference on uncertainty in artificial intelligence Morgan Kaufmann, 1995, pp. 338-345.

[38] J. Kazmierska, J. Malicki, "Apllication of the Navie Bayeian classifier to optimize treatment decisions," Journal of Radiotherapy and Oncology, 86(2), 2008, pp. 211-216.

[39] X.R. Marcela, J.M.T. Agma, T.Caetano, M.A.M. Paulo, "An Association Rule-Based Method to Support Medical Image Diagnosis With Efficiency," IEEE transactions on multimedia, 10( 2):2008, pp. 277-285.

## AUTHORS PROFILE

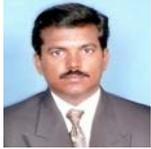

P.Rajendran obtained his MCA degree from Bharathidhasan University in 2000, ME Degree in Computer science and engineering from Anna University, Chennai, in 2005. He has started his teaching profession in the year 2000 in Vinayakamissions engineering college, salem. At present he is an Assistant Professor in department of computer science and engineering in K.S.Rangasamy college of Technology, Thiruchengode. . He has published 10 research papers in International and National Journals as well as conferences. He is a part time Ph.D research scalar in Anna University Chennai. His areas of interest are Data mining, Image mining and Image processing. He is a life member of ISTE.

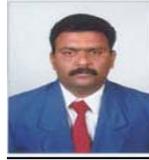

Dr. M. Madheswaran has obtained his Ph.D. degree in Electronics Engineering from Institute of Technology, Banaras Hindu University, Varanasi in 1999 and M.E degree in Microwave Engineering from Birla Institute of Technology, Ranchi, India. He has started his teaching profession in the year 1991 to serve his parent Institution Mohd. Sathak Engineering College, Kilakarai where he obtained his Bachelor Degree in ECE. He has served KSR college of Technology from 1999 to 2001 and PSNA College of Engineering and Technology, Dindigul from 2001 to 2006. He has been awarded Young Scientist Fellowship by the Tamil Nadu State Council for Science and Technology and Senior Research Fellowship by Council for Scientific and Industrial Research, New Delhi in the year 1994 and 1996 respectively. His research project entitled "Analysis and simulation of OEIC receivers for tera optical networks" has been funded by the SERC Division, Department of Science and Technology, Ministry of Information Technology under the Fast track proposal for Young Scientist in 2004. He has published 120 research papers in International and National Journals as well as conferences. He has been the IEEE student branch counselor at Mohamed Sathak Engineering College, Kilakarai during 1993-1998 and PSNA College of Engineering and Technology, Dindigul during 2003-2006. He has been awarded Best Citizen of India award in the year 2005 and his name is included in the Marquis Who's Who in Science and Engineering, 2006-2007 which distinguishes him as one of the leading professionals in the world. His field of interest includes semiconductor devices, microwave electronics, optoelectronics and signal processing. He is a member of IEEE, SPIE, IETE, ISTE, VLSI Society of India and Institution of Engineers (India).